\documentclass{article}
\usepackage{ijcai26}

\usepackage{times}
\usepackage{soul}
\usepackage{url}
\usepackage[hidelinks]{hyperref}
\usepackage[utf8]{inputenc}
\usepackage[small]{caption}
\usepackage{graphicx}
\usepackage{amsmath}
\usepackage{amsthm}
\usepackage{booktabs}
\usepackage{algorithm}
\usepackage{algorithmic}
\usepackage[switch]{lineno}
\usepackage{multirow}
\usepackage{xcolor}

\usepackage{subcaption} 
\usepackage{caption}

\usepackage[table]{xcolor}
\usepackage{booktabs}
\usepackage{graphicx}
\usepackage{pifont}
\usepackage{amsmath}
\usepackage{colortbl}

\usepackage{listings}
\usepackage{xcolor}

\definecolor{green!45!yellow!21}{HTML}{9ACD32}
\definecolor{green!50!black}{HTML}{008000}
\definecolor{OxfordBlue!10}{HTML}{E6EAF0}
\definecolor{OxfordBlue!20}{HTML}{CCD5E1}
\definecolor{OxfordBlue!30}{HTML}{B3C0D2}
\definecolor{AuroraBlue!100}{HTML}{7FB2E5}

\usepackage{amssymb}

\definecolor{TableHead}{RGB}{230, 240, 250} 
\definecolor{SectionColor}{RGB}{245, 245, 245} 
\definecolor{HighlightColor}{RGB}{255, 248, 220} 
\definecolor{DarkBlue}{RGB}{0, 51, 102}

\newcommand{\cmark}{\textcolor{green!50!black}{\ding{51}}}
\newcommand{\xmark}{\textcolor{red}{\ding{55}}}

\title{\textsc{DataCross}: A Unified Benchmark and Agent Framework for Cross-Modal Heterogeneous Data Analysis}

\author{
    Ruyi Qi$^{1,}$\footnote{These authors contributed equally to this work.} \and
    Zhou Liu$^{1,}$\footnotemark[1] \and
    Wentao Zhang$^{1,}$\thanks{Corresponding author: wentao.zhang@pku.edu.cn}
    \affiliations
    $^1$Peking University\\
    \emails
    qrymaple@gmail.com, liuzhoul82@gmail.com,
    wentao.zhang@pku.edu.cn
}

\begin{document}

\maketitle

\begin{abstract}
In real-world data science and enterprise decision-making, critical information is often fragmented across directly queryable structured sources (e.g., SQL, CSV) and ``zombie data" locked in unstructured visual documents (e.g., scanned reports, invoice images). Existing data analytics agents are predominantly limited to processing structured data, failing to activate and correlate this high-value visual information, thus creating a significant gap with industrial needs. To bridge this gap, we introduce DataCross, a novel benchmark and collaborative agent framework for unified, insight-driven analysis across heterogeneous data modalities. \textbf{DataCrossBench} comprises 200 end-to-end analysis tasks across finance, healthcare, and other domains. It is constructed via a human-in-the-loop reverse-synthesis pipeline, ensuring realistic complexity, cross-source dependency, and verifiable ground truth. The benchmark categorizes tasks into three difficulty tiers to evaluate agents’ capabilities in visual table extraction, cross-modal alignment, and multi-step joint reasoning.We also propose the \textbf{DataCrossAgent} framework, inspired by the ``divide-and-conquer" workflow of human analysts. It employs specialized sub-agents, each an expert on a specific data source, which are coordinated via a structured workflow of Intra-source Deep Exploration, Key Source Identification, and Contextual Cross-pollination. A novel reReAct mechanism enables robust code generation and debugging for factual verification. 
Experimental results show that DataCrossAgent achieves a 29.7\% improvement in factuality over GPT-4o and exhibits superior robustness on high-difficulty tasks, effectively activating fragmented ``zombie data" for insightful, cross-modal analysis.
\end{abstract}

\section{Introduction}

\label{sec:intro}
Large Language Models (LLMs) and multimodal foundation models have recently shown strong capabilities in reasoning, code generation, and tool use, enabling a new paradigm of \emph{data analysis agents} that can plan, execute Python/SQL, and produce analytical narratives \cite{deepseek2025r1,openai2025gpt5,yang2025qwen3}.
In real organizations, however, high-value evidence for decision-making is rarely stored in a single clean table.
Instead, it is fragmented across directly queryable structured systems (e.g., SQL/CSV/JSON) and a large amount of \emph{unstructured visual documents} (e.g., scanned reports, invoice images, screenshots of legacy systems) that often contain critical tables and records.
Such visually locked information is frequently `alive' for business operations but `dead' for analytics pipelines, and is therefore often referred to as \emph{zombie data}.

Despite progress, existing data analysis agents and benchmarks still fall short of what industrial analytics requires.
First, most agents assume the key evidence is already in structured formats, and thus cannot reliably \emph{activate zombie data}-i.e., extract tabular structure and values from pixels and treat them as queryable data.
Second, even when OCR or table extraction is available, performing \emph{cross-modal alignment} remains difficult: entities, column names, units, and identifiers shown in images must be normalized and matched to structured records (often with noise, abbreviations, or inconsistent formats).
Third, current evaluation protocols do not adequately measure end-to-end capability in this setting.
Text-to-SQL or single-table analysis benchmarks capture only part of the workflow, while many multimodal QA/RAG benchmarks emphasize retrieval over documents rather than \emph{joint analytical reasoning} that integrates multiple heterogeneous sources into verifiable insights.
As a result, we lack a reproducible way to evaluate whether an agent can (i) extract tables from images, (ii) align them with structured sources, and (iii) produce evidence-grounded analytical conclusions.

To address this gap, we introduce \textbf{DataCrossBench}, a benchmark for unified, insight-driven analysis across heterogeneous modalities. DataCrossBench consists of 200 end-to-end tasks spanning multiple domains such as finance and healthcare. Each task requires cross-source evidence gathering and joint reasoning, where key facts are distributed across multiple structured files and, in many tasks, tables embedded in images. Crucially, the benchmark is constructed with a human-in-the-loop \emph{reverse-synthesis} pipeline: we start from expert-validated analytical goals and target insights, then programmatically synthesize the underlying structured artifacts and corresponding visual documents so that conclusions are derivable and verifiable by execution. This design aims to make cross-modal analytics not only challenging and realistic, but also \emph{measurable} with traceable ground truth.

On top of the benchmark, we propose \textbf{DataCrossAgent}, a collaborative agent framework inspired by the divide-and-conquer workflow of human analysts. Rather than forcing a single agent to process all sources monolithically, DataCrossAgent assigns specialized sub-agents to different data sources and coordinates them to (a) deeply profile each source, including extracting table content from images, (b) identify key sources that are most informative for the given goal, and (c) generate and verify cross-source hypotheses through executable analysis. To improve robustness in tool use, we further employ a recursive Reasoning--Act mechanism (reReAct) that iteratively generates, debugs, and validates analysis code, grounding final claims in execution results and explicit evidence links.

Overall, this work contributes a new benchmark and a corresponding agent framework for cross-modal heterogeneous data analysis: 
\begin{itemize}
    \item DataCrossBench: We introduce a new benchmark for cross-modal, heterogeneous data analysis that makes structured--visual joint analytics explicit, verifiable, and scalable, filling a key gap in existing evaluations.
    \item DataCrossAgent: We propose a corresponding agent framework that activates zombie visual data and produces evidence-grounded insights via coordinated reasoning over multiple data sources.
\end{itemize}

\section{Related Work}
\label{sec:related_work} 

\subsection{Insight Discovery and Analytics}

Benchmarks for data science and analytics have evolved from single-source, single-turn question answering to end-to-end workflows that require planning, tool use, and open-ended reporting.
Early benchmarks primarily target semantic parsing over structured data, such as Text-to-SQL and table QA, where success is defined by a specific query result or short answer (e.g., Spider, WikiTableQuestions).
More recent benchmarks move toward executable analysis by evaluating code generation and debugging over tabular datasets, emphasizing Python/SQL tool use and multi-step reasoning (e.g., DS-1000 and follow-up data-analysis evaluations) \cite{ds1000,dseval,text2analysis,dabench}.

In parallel, another line of work focuses on \emph{insight discovery} rather than factoid answering.
InsightBench and AgentPoirot evaluate whether models can synthesize higher-level findings and narratives under an analysis goal, better reflecting real analytical practice \cite{insightbench}.
However, these settings typically assume a single structured table, leaving cross-file integration and heterogeneous evidence underexplored.
Some benchmarks begin to incorporate multiple sources or multiple structured types \cite{dabstep,unidatabench}, yet they largely exclude unstructured visual documents that often contain critical business evidence (e.g., scanned reports, screenshots, and invoice images).

Meanwhile, document-centric multimodal benchmarks have advanced rapidly in the vision--language community.
WikiMixQA evaluates multimodal reasoning over \emph{tables and charts} in long Wikipedia documents, emphasizing long-context retrieval and cross-modal synthesis \cite{foroutan2025wikimixqa}.
UniDoc-Bench formalizes \emph{multimodal RAG} evaluation over a large PDF corpus, enabling unified comparisons across text-only, image-only, and fused retrieval \cite{peng2026unidocbench}.
Domain-specific benchmarks such as MedInsightBench further study goal-driven, multi-step \emph{medical insight discovery} from pathology images and reports \cite{zhu2025medinsightbench}.
Complementarily, Chart-RVR suggests that verifiable surrogate objectives (e.g., chart-type prediction and chart-to-table reconstruction) and process-level constraints improve robust and explainable chart reasoning \cite{sinha2025chartrvr}, highlighting structured extraction from visual artifacts as a prerequisite for reliable analytics.

Despite these advances, a critical gap remains for evaluating data science agents in realistic settings: high-value information is often distributed across \emph{heterogeneous} sources, where some are directly queryable (SQL/CSV/JSON) while others exist only as \emph{unstructured visual documents} containing tables.
To bridge this gap, we introduce DataCrossBench, which explicitly evaluates cross-source retrieval, cross-modal alignment, and joint reasoning over structured data and visual tables, with verifiable evidence links for each reported insight.

\subsection{Data Analysis Agents}
Data analysis agents have evolved from code-generation assistants to systems that support autonomous planning and increasingly multimodal decision-making. Early paradigms primarily used LLMs as translators from intent to executable Python/SQL within interactive execution loops. More recent work adopts agentic architectures that couple iterative reasoning with tool use to cover end-to-end workflows (cleaning, EDA, modeling, visualization), often via multi-agent collaboration frameworks such as Microsoft’s AutoGen and related ``agentic AI” design patterns discussed in industry retrospectives. In parallel, practical systems are moving toward higher automation in the data-science lifecycle, e.g., Google Cloud’s product direction on agents for data teams.

Multimodality has become a central capability for real deployments, reflecting the push toward assistants that can operate over mixed media. On the research side, MMCTAgent \cite{kumar2024mmctagent} demonstrates an agentic planner–critic style for structured reasoning over large visual corpora. Despite these advances, agents still struggle with cross-system analysis across heterogeneous formats and granular logs \cite{just2024data}, and frequently exhibit a ``silo effect”: they process sources sequentially or rely on shallow schema concatenation, limiting their ability to uncover latent links across tables, text, and visual artifacts. DataCrossAgent addresses this gap via Contextual Cross-pollination, where specialized sub-agents reinterpret one source in the context of another to enable cross-source and cross-modal insight discovery that single-stream pipelines often miss.

\begin{figure*}[htbp]
    \centering
    \begin{subfigure}[t]{0.33\textwidth}
        \centering
        \includegraphics[width=\linewidth]{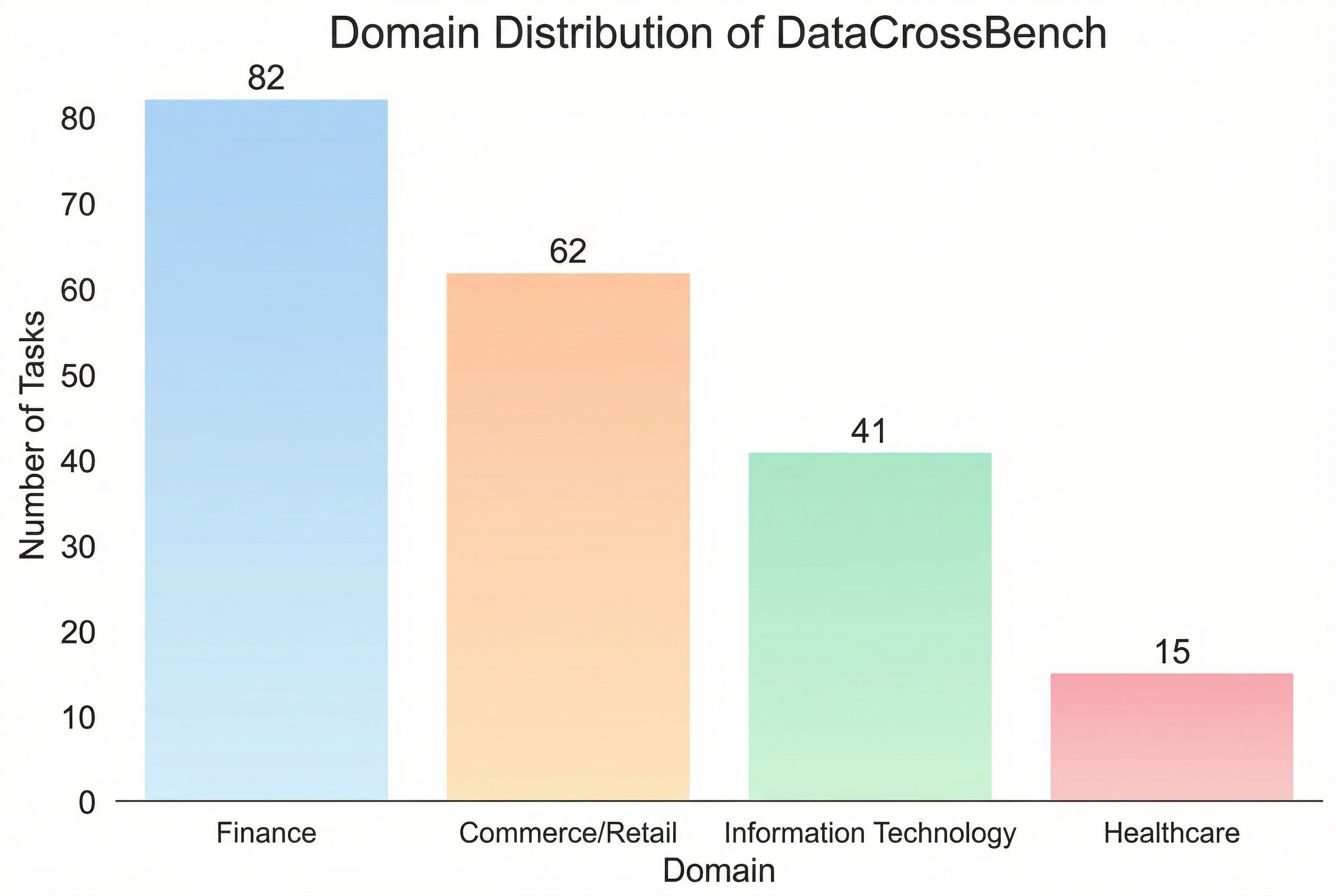}
        \caption{\textbf{Domain distribution} of DataCrossBench across finance, healthcare, information technology, and commerce/retail.}
        \label{fig:domain_distribution}
    \end{subfigure}\hfill
    \begin{subfigure}[t]{0.33\textwidth}
        \centering
        \includegraphics[width=\linewidth]{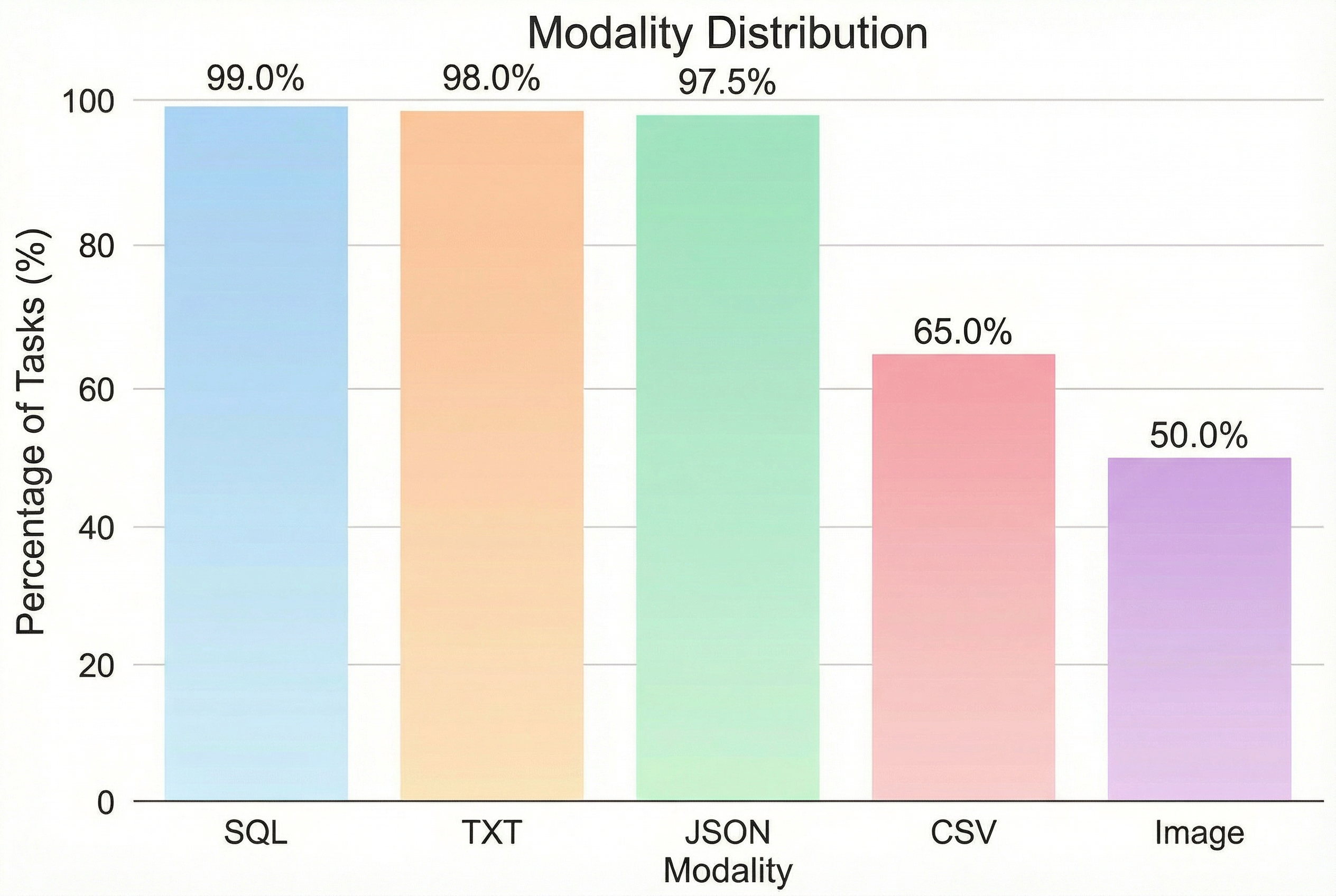}
        \caption{\textbf{Modality distribution} of DataCrossBench. The figure shows the proportion of tasks involving each data modality (CSV, SQL, JSON, TXT, Image). A task can include multiple modalities.}
        \label{fig:modality_distribution}
    \end{subfigure}\hfill
    \begin{subfigure}[t]{0.20\textwidth}
        \centering
        \includegraphics[width=\linewidth]{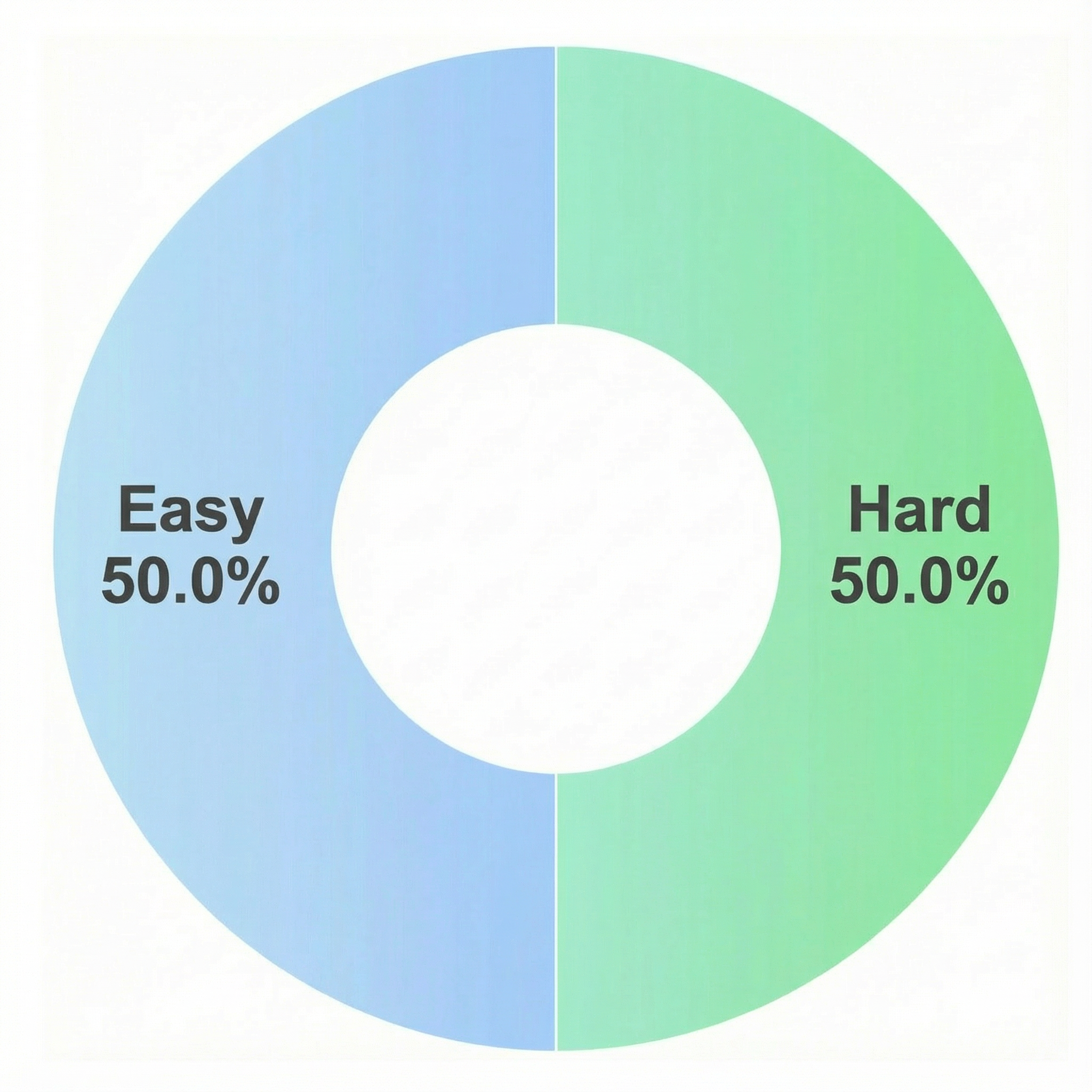}
        \caption{\textbf{Difficulty tier breakdown} of the 200 tasks in DataCrossBench (Easy, Hard), defined by modality composition.}
        \label{fig:difficulty_tiers}
    \end{subfigure}

    \caption{\textbf{DataCrossBench distributions.} (a) Domain distribution. (b) Modality distribution. (c) Difficulty tier breakdown.}
    \label{fig:datacrossbench_distributions}
\end{figure*}

\section{DataCrossBench: A Multi-Source Heterogeneous Benchmark}
\subsection{Design Philosophy and Comparison}

Despite rapid progress in data analysis agents, existing benchmarks remain constrained by idealized assumptions that emphasize a single modality. Classic datasets such as Spider focus primarily on Text to SQL, while more recent benchmarks such as DS-1000 introduce code generation but still operate on preprocessed CSV files, failing to capture the complexity of real settings characterized by heterogeneous sources, interconnectivity across sources, and semantic ambiguity. In practice, data extend beyond structured tables to include semi structured JSON and logs, as well as unstructured visual documents that often contain critical evidence, such as scanned PDF reports and screenshots. Moreover, valuable insights rarely come from a single file and instead require entity alignment across sources to establish a consistent evidence chain, while remaining robust to noise, redundancy, and inconsistency. Most existing benchmarks do not evaluate the ability to extract information from visual documents and jointly query it with structured databases. To address this gap, we propose DataCrossBench, a new benchmark designed to evaluate end to end data analysis in ecologically valid settings. As shown in Table \ref{Table:DatasetTable}, DataCrossBench provides a multi modal, noisy, cross file evaluation environment that tests whether agents can perform end to end analysis under realistic constraints.

\begin{table*}[th!]
\centering
\caption{Comparison of \textbf{DataCrossBench} with other benchmarks. We additionally highlight whether a benchmark involves \textbf{visual documents} (vision modality) that must be jointly analyzed with structured sources.}
\label{Table:DatasetTable}
\vspace{5pt}
\renewcommand{\arraystretch}{1.2}
\setlength{\tabcolsep}{6pt}
\scalebox{0.78}{
\begin{tabular}{l l c c c c c}
\toprule[1.5pt]
\rowcolor{TableHead}
\textbf{Dataset} & \textbf{Analysis Source} & \textbf{Multi-source} & \textbf{Multi-modal (Vision)} & \textbf{Task Size} & \textbf{File Size} & \textbf{Rows/File} \\
\midrule

\rowcolor{SectionColor}
\multicolumn{7}{l}{\textbf{\textit{Specific Data Analysis}} (Specific answer for a given question)} \\
\midrule
DS-1000 \cite{ds1000} & StackOverflow & \xmark{} (data.frame) & \xmark{} & 1000 & -- & -- \\
DSEval \cite{dseval} & LLM Gen. \& Human & \xmark{} (csv) & \xmark{} & 825 & 44 & 23,630 \\
Text2Analysis \cite{text2analysis} & LLM Gen. \& Human & \xmark{} (csv) & \xmark{} & 2249 & 347 & -- \\
DABench \cite{dabench} & LLM Gen. \& Human & \xmark{} (csv) & \xmark{} & 257 & 52 & 2,024 \\
DABStep \cite{dabstep} & Real-world Data & \cmark{} (3 types) & \xmark{} & 450 & 7 & 3,121 \\

\midrule
\rowcolor{SectionColor}
\multicolumn{7}{l}{\textbf{\textit{Multimodal QA / RAG}} (Specific answer over visual documents)} \\
\midrule
WikiMixQA \cite{foroutan2025wikimixqa} & Wikipedia pages (tables \& charts) & \cmark{} (multi-page doc) & \cmark{} & 1000 & 4000 & -- \\
UniDoc-Bench \cite{peng2026unidocbench} & PDF corpus (page-level retrieval) & \cmark{} (multi-doc/pages) & \cmark{} & 1600 & 70000 & -- \\

\midrule
\rowcolor{SectionColor}
\multicolumn{7}{l}{\textbf{\textit{High-level Data Analysis}} (Insights extracted for open-ended goals)} \\
\midrule
InsightBench \cite{insightbench} & LLM \& Human & \xmark{} (csv) & \xmark{} & 100 & 112 & 545 \\
UniDataBench \cite{unidatabench} & Real-world Data & \cmark{} (Multi-type) & \xmark{} & 100 & 223 & 17,945 \\
MedInsightBench \cite{zhu2025medinsightbench} & Medical cases (multimodal) & \xmark{} (single case) & \cmark{} & 332 & -- & -- \\
\midrule

\rowcolor{HighlightColor}
\textbf{DataCrossBench (Ours)} & \textbf{LLM \& Human} & \textbf{\cmark{} (4+ types)} & \textbf{\cmark{}} & \textbf{200} & \textbf{846} & \textbf{11,002} \\
\bottomrule[1.5pt]
\end{tabular}
}
\end{table*}

\subsection{Data Construction}
We employ a human-in-the-loop \textit{reverse-synthesis} pipeline to ensure the data is both realistic and challenging.
First, we curate authoritative reports (PDFs) from four domains (finance, healthcare, information technology, and commerce/retail) as seed documents, and utilize an LLM to extract high-level analytical goals and ground-truth insights, which are rigorously verified by human experts.
Next, guided by these validated insights, we collaborate with a Code LLM to programmatically synthesize the underlying raw data files (e.g., SQL, CSV tables) required to derive the target conclusions. During this phase, we enforce strict complexity constraints---such as requiring specific file counts and ensuring large-scale data volume (e.g., $>3,000$ records per file)---to simulate real-world data magnitude.
Finally, the pipeline undergoes an iterative validation process: human annotators perform execution checks to identify and repair any consistency errors or crashes, followed by a definitive quality assurance review to guarantee that the synthesized multi-source data logically and accurately supports the ground-truth insights.
After all structured data artifacts are completed, we then produce auxiliary unstructured and multimodal resources—such as writing accompanying \texttt{.txt} files and generating visualization charts—as additional heterogeneous data sources.

\subsection{Dataset Statistics and Diversity}
DataCrossBench comprises a total of 200 tasks, each designed as an end-to-end data analysis problem requiring cross-source evidence gathering and reasoning. To ensure broad coverage and mitigate domain-specific bias, the benchmark spans multiple real-world sectors, including finance, healthcare, information technology, and commerce and retail, as shown in Figure~\ref{fig:domain_distribution}. The dataset incorporates heterogeneous data modalities commonly observed in practice, involving \textit{CSV files, SQL databases, JSON objects, TXT reports, and images} such as scanned tables or invoices (see Figure~\ref{fig:modality_distribution}). Based on whether visual documents are involved, we categorize tasks into two difficulty tiers:
\textit{Easy} tasks involve only structured or textual sources (e.g., CSV/SQL/JSON/TXT) and focus on cross-file reasoning without any image inputs;
\textit{Hard} tasks require parsing and integrating visual documents (i.e., include at least one image), demanding visual table extraction, normalization, and cross-modal alignment with structured sources to support multi-step verification across data sources.

\subsection{Quality Control and Human Verification}
To ensure the robustness and accuracy of DataCrossBench, we implemented a rigorous two-stage quality control framework comprising automated sanity checks and expert human verification. Initially, an automated pipeline verifies the executability of all reference SQL and Python code and checks for data file corruption. Subsequently, we engaged annotators with professional backgrounds in data science to perform a double-blind review of the tasks. This manual verification process adheres to three strict criteria: (1) \textit{Cross-source Necessity}, confirming that questions strictly require evidence gathering across multiple sources and cannot be answered by relying on a single file; (2) \textit{Logic Chain Validation}, ensuring the reasoning path constitutes a unique and reasonable evidence chain; and (3) \textit{Answer Consistency}, guaranteeing that the ground truth aligns perfectly with manual calculations. To facilitate this workflow, we developed a dedicated human verification interface. Further details regarding the specific protocols and annotation guidelines are provided in supplementary material.

\subsection{Evaluation Metrics} 
\label{sec:Evaluation Metrics}
DataCrossBench tasks demand open-ended analytical reports and complex cross-source reasoning, which pose significant challenges for standard evaluation. While prior research has validated the feasibility of utilizing Large Language Models (LLMs) as evaluators for free-form analytical outputs, pure LLM-based scoring often suffers from \textbf{central tendency bias}, where the model leans toward assigning median scores and fails to capture nuanced quality differences. To mitigate this, we propose a comprehensive multi-dimensional scoring framework that integrates deterministic heuristic metrics with LLM-based semantic assessment. This hybrid approach ensures a balanced evaluation across four key dimensions: \textbf{Factuality}, \textbf{Completeness}, \textbf{Logic}, and \textbf{Insightfulness}, thereby providing both the rigor of hard data verification and the depth of semantic understanding. The final score $S_{total}$ is a weighted average of four key dimensions: \textbf{Factuality} ($w_f=0.30$), \textbf{Completeness} ($w_c=0.25$), \textbf{Logic} ($w_l=0.20$), and \textbf{Insightfulness} ($w_i=0.25$), where $w_f$ is set a bit higher to emphasize numerical accuracy.

\begin{equation}
S_{4-dim} = w_f \cdot S_{factu} + w_c \cdot S_{comp} + w_l \cdot S_{logic} + w_i \cdot S_{ins}
\end{equation}

\vspace{2pt}
\paragraph{Factuality ($S_{fact}$)}
Factuality measures the alignment of numerical data and entity descriptions between the predicted insights ($I_{pred}$) and the ground truth ($I_{gt}$). This metric is composed of two equally weighted sub-components:

\begin{itemize}
    \item \textit{\textbf{Numerical Consistency ($S_{num}$):}} A deterministic metric where we extract all numerical values and units. The score is defined as:
    \begin{equation}
    S_{factu\_num} = \frac{|N_{pred} \cap N_{gt}|}{|N_{gt}|}
    \end{equation}
    where $N$ denotes the set of extracted numerical entities.
    \item \textit{\textbf{Semantic Factuality ($S_{factu\_llm}$):}} An LLM-based evaluation using a customized prompt to assess the correctness of key entities and temporal trends on a scale of 1--10.
\end{itemize}

The final factuality score is $S_{factu} = 0.5 \cdot S_{factu\_num} + 0.5 \cdot (S_{factu\_llm} / 10)$.

\vspace{3pt}
\paragraph{Completeness ($S_{comp}$)}
Completeness assesses the information coverage. We employ an embedding-based approach to measure the semantic overlap. For each ground-truth insight $g \in I_{gt}$, we find its maximum cosine similarity across all predicted insights $p \in I_{pred}$:
\begin{equation}
S_{comp} = \frac{1}{|I_{gt}|} \sum_{g_i \in I_{gt}} \max_{p_j \in I_{pred}} \text{cos}(\mathbf{e}_{g_i}, \mathbf{e}_{p_j})
\end{equation}
where $\mathbf{e}$ represents the semantic embedding vector.

\vspace{3pt}
\paragraph{Logic ($S_{logic}$)}
The logic dimension evaluates the internal coherence and reasoning quality. The LLM assesses: \textit{\textbf{Attribution: }}validity of causal relationships; \textit{\textbf{Trend Inference}: }evidence-based extrapolation; \textit{\textbf{Comparison}: }objectivity of analysis. The score $L \in [1, 10]$ is normalized as $S_{logic} = L/10$.

\vspace{3pt}
\paragraph{Insightfulness ($S_{ins}$)}
Insightfulness measures the value-add of the generated content. The LLM evaluates non-triviality, actionability, and novelty. To ensure stability, we adopt a G-Eval style approach, calculating a weighted average of the top-5 log-probabilities of the output tokens.

\section{DataCrossAgent: A Corresponding Agent Framework}

\begin{figure*}[t]
    \centering
    \includegraphics[width=1.0\textwidth]{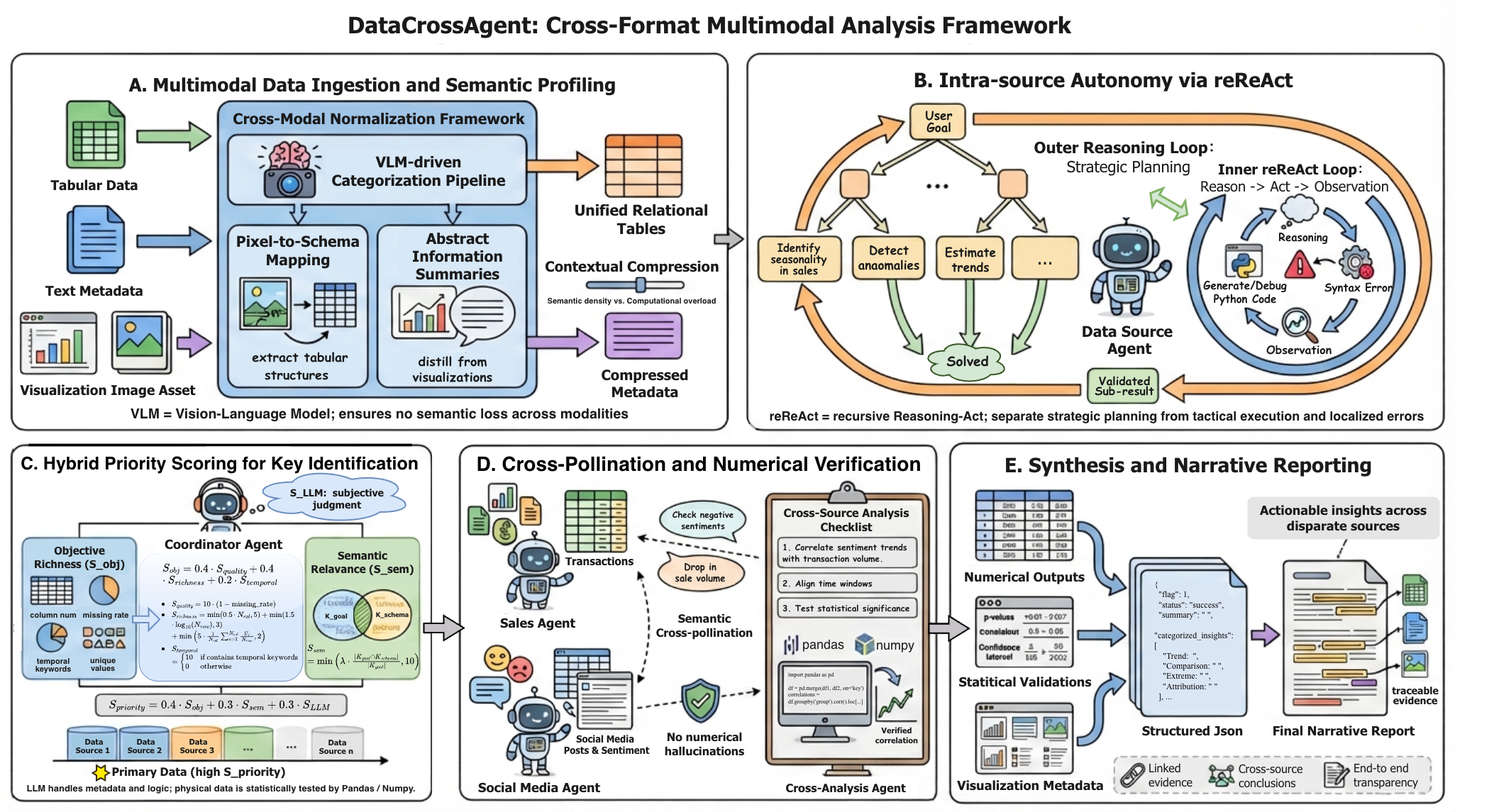}
    \caption{\textbf{The overall architecture of the DataCross framework.} A multi-agent orchestration to integrate heterogeneous data sources.}
    \label{fig:framework_architecture}
\end{figure*}

The DataCross framework addresses the fragmentation of cross-format data analysis through a multi-agent orchestration that mimics human cognitive ``divide-and-conquer" heuristics. Unlike traditional single-agent systems that struggle with context window limits and numerical hallucinations, DataCross employs a stratified architecture designed to solve the challenges of multimodal integration and reasoning.

\subsection{Multimodal Data Ingestion and Semantic Profiling}

To resolve the challenge of non-unified data entry points, we introduce a \textbf{Cross-Modal Normalization Framework}. This framework leverages a VLM-driven pipeline to perform fine-grained asset identification. For visual inputs, we adopt a ``Classify-then-Extract'' strategy to handle different image types: \textbf{tabular assets} are processed by \texttt{read\_image()}, which maps visual grids into structured \texttt{DataFrames} to facilitate precise calculation; \textbf{visualizations} (e.g., charts) are processed via \texttt{generate\_chart\_description()}, synthesizing semantic summaries as auxiliary context. Furthermore, the ingestion of textual metadata is optimized through a \textbf{contextual compression process}, balancing the trade-off between semantic density and computational overhead.

\subsection{Intra-source Autonomy via reReAct}

To ensure deep exploration without losing sight of the global objective, each agent operates within a \textbf{recursive Reasoning-Act (reReAct)} structure. This dual-loop mechanism separates strategic planning from tactical execution. The outer reasoning layer decomposes the overarching task into a tree-like hierarchy of sub-problems (e.g., ``Identify seasonality in sales"). For each leaf node, an inner loop (Reasoning $\rightarrow$ Act $\rightarrow$ Observation) is triggered to generate and iteratively debug Python code. This stratified approach ensures that syntax errors or data anomalies do not derail the entire analysis, as the agent remains localized to the sub-problem until a valid observation is retrieved.

\subsection{Hybrid Priority Scoring for Key Source Identification}

In scenarios with high data dimensionality, treating all sources with equal weight leads to computational inefficiency and ``noise" in the reasoning process. To address this, we introduce a \textbf{Hybrid Priority Scoring} mechanism that quantifies the value of each data source $i$ through a tripartite evaluation of objective quality, semantic relevance, and subjective utility.

\paragraph{Objective Richness ($S_{obj}$).} This score measures the intrinsic information density of the dataset. Let $D$ be a dataset with rows $R$ and columns $C$. The objective score is defined as:
\begin{equation}
    S_{obj} = 0.4 \cdot S_{quality} + 0.4 \cdot S_{richness} + 0.2 \cdot S_{temp}
\end{equation}
where $S_{quality} = 10 \cdot (1 - \text{missing\_rate})$, and $S_{richness}$ is a composite of column count $|C|$, logarithmic row scaling $\log_{10}(|R|)$, and the average unique value ratio $\bar{\rho}_{unique}$ across all dimensions. $S_{temp}$ is a binary indicator ($0$ or $10$) based on the presence of temporal keywords, prioritizing datasets that allow for longitudinal analysis.

\paragraph{Semantic Relevance ($S_{sem}$).} To ensure the agent focuses on data pertinent to the user's goal, we calculate the overlap between the goal's keyword set $K_{goal}$ and the data schema's keyword set $K_{schema}$, where $\lambda$ is a scaling factor:
\begin{equation}
    S_{sem} = \min\left( \lambda \cdot \frac{|K_{goal} \cap K_{schema}|}{|K_{goal}|}, 10 \right)
\end{equation}

\paragraph{Total Priority Score.} The final importance rank $S_{priority}$ is a weighted aggregation:
\begin{equation}
    S_{priority} = 0.4 \cdot S_{obj} + 0.3 \cdot S_{sem} + 0.3 \cdot S_{LLM}
\end{equation}
where $S_{LLM}$ represents a high-level subjective judgment from the Coordinator Agent. This score allows the system to identify ``Primary Data" (high-score pivots) and ``Auxiliary Data", ensuring that cross-source hypotheses are generated with a clear hierarchy of evidence.

\subsection{Cross-Pollination and Numerical Verification}

The core of the framework lies in ``Semantic Cross-pollination", where agents view external datasets through the lens of their own expertise. A Sales Agent, for instance, might analyze the summary of a Social Media Agent to hypothesize correlations between sentiment trends and transaction volume. These hypotheses are formalized into a ``Cross-Source Analysis Checklist".

To prevent the numerical hallucinations common in LLMs, the actual ``joining" of data is performed by a specialized \textbf{Cross-Analysis Agent}, which generates executable Python code (utilizing Pandas and NumPy) to perform physical data merges and statistical tests in a sandboxed environment. This decoupled execution ensures high numerical fidelity, while enables the framework to handle larger-scale datasets more robustly.

\subsection{Synthesis and Narrative Reporting}

The final stage involves the convergence of findings into a coherent narrative. The system integrates numerical outputs, statistical validations, and visualization metadata into a structured JSON format, which is then synthesized into a final report. This report avoids fragmented reporting by linking evidence from disparate sources, providing actionable insights with full traceability.

\begin{table*}[t]
\centering
\caption{Four-dimension Score Performance}
\label{tab:four_dim_scores}
\vspace{4pt}
\renewcommand{\arraystretch}{1.08} 
\setlength{\tabcolsep}{3.2pt}     
\footnotesize                     
\resizebox{\textwidth}{!}{%
\begin{tabular}{l *{15}{c}}
\toprule[1.5pt]

\rowcolor{TableHead}
\multirow{3}{*}{\textbf{Method}} &
\multicolumn{15}{c}{\textbf{Four-dim Insight Score}} \\

\rowcolor{TableHead}
& \multicolumn{3}{c}{\textbf{Factuality}}
& \multicolumn{3}{c}{\textbf{Completeness}}
& \multicolumn{3}{c}{\textbf{Logic}}
& \multicolumn{3}{c}{\textbf{Insightfulness}}
& \multicolumn{3}{c}{\textbf{Overall}} \\

\rowcolor{TableHead}
& \textbf{Easy} & \textbf{Hard} & \textbf{Avg.}
& \textbf{Easy} & \textbf{Hard} & \textbf{Avg.}
& \textbf{Easy} & \textbf{Hard} & \textbf{Avg.}
& \textbf{Easy} & \textbf{Hard} & \textbf{Avg.}
& \textbf{Easy} & \textbf{Hard} & \textbf{Avg.} \\

\cmidrule(lr){2-4}\cmidrule(lr){5-7}\cmidrule(lr){8-10}\cmidrule(lr){11-13}\cmidrule(lr){14-16}

\rowcolor{SectionColor}
\multicolumn{16}{l}{\textbf{\textit{General LLMs (Closed-Source \& Open-Source)}}} \\
GPT-4o
& 0.3313 & 0.3269 & 0.3291
& 0.4689 & 0.4160 & 0.4425
& 0.4376 & 0.3815 & 0.4096
& 0.3835 & 0.3272 & 0.3554
& 0.4000 & 0.3602 & 0.3801 \\
Gemini-2.5-flash
& 0.2957 & 0.3198 & 0.3078
& 0.3672 & 0.3517 & 0.3594
& 0.4063 & 0.3902 & 0.3982
& 0.3930 & 0.3892 & 0.3911
& 0.3600 & 0.3592 & 0.3596 \\
Qwen3-vl
& 0.3074 & 0.1825 & 0.2473
& 0.4112 & 0.2328 & 0.3220
& 0.4337 & 0.2436 & 0.3398
& 0.4486 & 0.2680 & 0.3562
& 0.3939 & 0.2287 & 0.3117 \\

\midrule

\rowcolor{SectionColor}
\multicolumn{16}{l}{\textbf{\textit{Specialized Agent Frameworks}}} \\
AgentPoirot (GPT-4o)
& 0.3069 & 0.2391 & 0.2730
& 0.4409 & 0.3851 & 0.4130
& 0.5652 & \textbf{0.5683} & 0.5668
& \textbf{0.5517} & \textbf{0.5445} & \textbf{0.5481}
& 0.4533 & 0.4178 & 0.4356 \\

\rowcolor{HighlightColor}
\textbf{DataCross (GPT-4o)}
& \textbf{0.4508} & \textbf{0.4031} & \textbf{0.4270}
& \textbf{0.5720} & \textbf{0.5368} & \textbf{0.5544}
& \textbf{0.5912} & 0.5645 & \textbf{0.5779}
& 0.5462 & 0.5326 & 0.5394
& \textbf{0.5331} & \textbf{0.5012} & \textbf{0.5172} \\

\bottomrule[1.5pt]
\end{tabular}
}

\vspace{3pt}
\begin{flushleft}
\footnotesize \textit{Note: \textbf{Bold signals:} Bolded values indicate the best performance in each column. \textbf{Score range:} The Four-dim Insight Score is normalized to $[0, 1]$, where higher values represent better performance. \textbf{Level rule:}``Easy'' refers to tasks involving only structured or textual data, while ``Hard'' denotes challenges requiring the parsing and integration of visual document images (e.g., scanned tables and charts). }
\end{flushleft}

\end{table*}

\begin{table*}[t]
\centering
\begin{minipage}{0.5\textwidth}
    \centering
    \caption{Ablation Study Results on Four-dimension Insight Score}
    \label{tab:ablation_four_dim}
    \vspace{5pt}
    \renewcommand{\arraystretch}{1.2}
    \setlength{\tabcolsep}{3.5pt} 
    \scalebox{0.74}{ 
    \begin{tabular}{lccccc}
    \toprule[1.5pt]
    \rowcolor{TableHead}
    \multirow{2}{*}{\textbf{}} &
    \multicolumn{5}{c}{\textbf{Four-dim Insight Score}} \\
    \rowcolor{TableHead}
    & \textbf{Factuality} & \textbf{Completeness} & \textbf{Logic} & \textbf{Insightfulness} & \textbf{Overall} \\
    \midrule

    \rowcolor{HighlightColor}
    \textbf{Full DataCrossAgent}
    & \textbf{0.4268 $\pm$ 0.1493}
    & \textbf{0.5543 $\pm$ 0.0849}
    & \textbf{0.5778 $\pm$ 0.1085}
    & \textbf{0.5394 $\pm$ 0.0837}
    & \textbf{0.5170 $\pm$ 0.0655} \\
    \midrule

    \rowcolor{SectionColor}
    \multicolumn{6}{l}{\textbf{\textit{Ablation Experiments}}} \\
    w/o reReAct
    & 0.3407 $\pm$ 0.1597 & 0.5017 $\pm$ 0.0862 & 0.4886 $\pm$ 0.0916 & 0.4652 $\pm$ 0.0858 & 0.4417 $\pm$ 0.0719 \\
    w/o Key-identification
    & 0.3917 $\pm$ 0.1543 & 0.5385 $\pm$ 0.0835 & 0.5378 $\pm$ 0.1235 & 0.4956 $\pm$ 0.0936 & 0.4836 $\pm$ 0.0699 \\
    w/o Cross-pollination
    & 0.3908 $\pm$ 0.1563 & 0.5320 $\pm$ 0.0904 & 0.5357 $\pm$ 0.1100 & 0.4840 $\pm$ 0.0785 & 0.4784 $\pm$ 0.0691 \\
    w/o Key \& Cross
    & 0.3283 $\pm$ 0.1576 & 0.4691 $\pm$ 0.0943 & 0.4959 $\pm$ 0.0831 & 0.4354 $\pm$ 0.0806 & 0.4238 $\pm$ 0.0721 \\
    \bottomrule[1.5pt]
    \end{tabular}
    }
    \vspace{3pt}
    \begin{flushleft}
        \footnotesize \textit{Note: ``w/o'' indicates ``without''; Four-dim Insight Score ranges from 0 to 1, and higher values indicate better performance.}
    \end{flushleft}
\end{minipage}
\hfill
\begin{minipage}{0.28\textwidth}
    \centering
    \includegraphics[width=\textwidth]{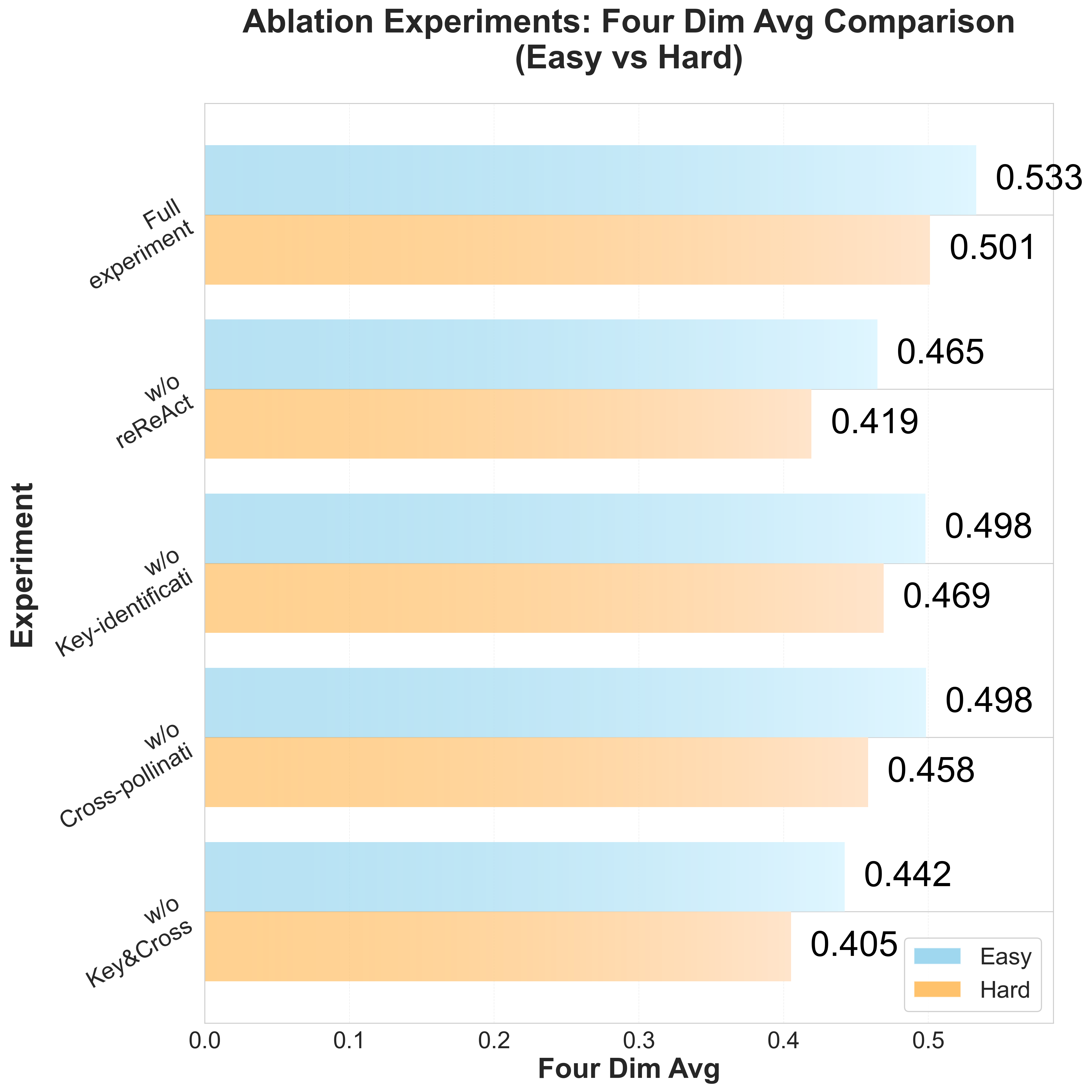}
    \captionof{figure}{Performance trends across different ablation and different levels.}
    \label{fig:ablation_score}
\end{minipage}
\end{table*}

\section{Experiments}
\label{sec:experiments}

\subsection{Setup}
To comprehensively validate DataCross's capability in handling heterogeneous data, we conduct a series of testing experiments on our benchmark.

\subsubsection{Baselines}
We evaluate the following baselines on the benchmark:

\paragraph{\textit{\textbf{Large Multimodal Models (LMMs):}}} Directly utilize several LMMs to generate insights, including GPT-4o \cite{gpt4o_system_card}, Qwen3-vl \cite{bai2025qwen25vl} and Gemini-2.5-flash \cite{gemini_flash}.
\paragraph{\textit{\textbf{AgentPoirot:}}} Given a dataset and an objective, this agent achieves end-to-end deep insight mining through continuous questioning. In our experiments, we organize various heterogeneous data into usable tabular formats as input and allow the agent to iteratively generate answers.
\paragraph{\textit{\textbf{DataCrossAgent:}}} Our proposed agent system discovers high-quality insights through more focused iterative cycles and cross-agent interaction design.

\subsubsection{Agent Deployment Details}
In each agent framework, we use GPT-4o as our backbone LMMs. All LMMs are configured with a temperature of 0 to ensure deterministic outputs. During operation, agents are set to explore questions for 2 iterative rounds, generating 2 new questions per round to ensure comprehensiveness.

\subsubsection{Evaluation Setup}
We evaluate the performance of our proposed Agent using the multi-dimensional benchmark and metrics defined in Section \ref{sec:Evaluation Metrics}. Specifically, we compute the weighted total score $S_{total}$ across all test cases to compare our method against baselines. For LLM-based metrics ($S_{factu\_llm}, S_{logic}, S_{ins}$), we utilize GPT-4o as the evaluator to ensure high-quality and consistent scoring, with the log-probability calculation applied to the \textit{Insightfulness} dimension for enhanced granularity.

\subsection{Main Results}
Table~\ref{tab:four_dim_scores} evaluates DataCross against various baselines, revealing the distinct advantages of both our benchmark design and the agent framework.

\paragraph{Benchmark Discriminative Power.}DataCrossBench exhibits strong \textbf{discriminative power} through its stratified Easy/Hard task design. All evaluated models show a consistent performance decay ($\Delta$) in Hard scenarios. Specifically, vanilla LMMs experience a significant ``Logic Gap", with overall scores dropping by over $10\%$. This validates that the Hard set successfully introduces high-entropy reasoning challenges that exceed simple pattern matching, effectively exposing the limitations of current models.

\paragraph{Agent Sota Performance.}
DataCrossAgent achieves \textbf{state-of-the-art (SOTA) results}, reaching a peak Overall score of $0.5172$. A key highlight is its \textbf{Factuality}, which outperforms the strongest baseline (GPT-4o) by a relative \textbf{29.7\%}, showing effectiveness at suppressing numerical hallucinations than direct generation approaches.

\paragraph{Robustness and Stability.}Beyond peak scores, DataCross demonstrates the \textbf{stability} across varying task complexities. While other models' performance collapses on Hard tasks (e.g., Qwen3-vl’s performance drops by $16.52\%$ in absolute terms), DataCross maintains a remarkably low decay rate ($\Delta = -0.0319$). This robustness proves it essential for navigating complex, multi-source environments, preventing the agent from becoming overwhelmed by data ``noise".

\subsection{Ablation Experiments}
To quantify the contribution of each module, we conducted an ablation study as shown in Table~\ref{tab:ablation_four_dim}.

\paragraph{Effect of reReAct.}
Removing the \textbf{reReAct} mechanism (\textit{w/o reReAct}) results in the most dramatic performance decline, with the Overall score dropping from $0.5170$ to $0.4417$. Most notably, \textit{Logic} and \textit{Factuality} decrease by $15.4\%$ and $20.1\%$ respectively. This underscores that the dual-loop structure—separating strategic decomposition from tactical code debugging—is the backbone of reasoning stability. Without this, the system is prone to cascading errors where a single syntax failure or misinterpreted data row derails the entire analysis.

\paragraph{Effect of Key Source Identification and Cross-Pollination.}
The exclusion of Key-identification and Cross-pollination (\textit{w/o Key \& Cross}) reduces the agent's performance to a level near vanilla GPT-4o ($0.4238$ vs $0.3801$).
\begin{itemize}
\item \textbf{Key-identification} acts as a precision filter; without it, the agent's attention is fragmented, leading to lower \textbf{Completeness}.
\item \textbf{Cross-pollination} is essential for \textbf{Insightfulness}. When removed, the agent primarily generates localized insights from single sources, failing to discover the non-trivial patterns that emerge only when disparate data sources are synthesized (e.g., correlating visual dashboard trends with structured sales logs).
\end{itemize}

\section{Conclusion}
\label{sec:conclusion}

In this work, we address a critical gap in data analytics agent research: the inability to unify structured data analysis with the activation of "zombie data" locked in visual documents. We introduce DataCrossBench, a heterogeneous benchmark with 200 end-to-end tasks spanning multiple domains, constructed via a human-in-the-loop reverse-synthesis pipeline to ensure realistic complexity and verifiable ground truth. Complementarily, we propose DataCrossAgent, a collaborative framework that leverages specialized sub-agents, reReAct recursive reasoning, and cross-source pollination to enable robust cross-modal alignment and joint reasoning. Experimental results demonstrate that DataCrossAgent outperforms state-of-the-art baselines across key dimensions (Factuality, Completeness, Logic, Insightfulness), particularly in suppressing numerical hallucinations and maintaining robustness on hard multi-modal tasks. This work advances the frontier of data analytics agents from structured-data exclusivity to true global data governance, providing a practical foundation for industrial-grade heterogeneous data analysis.
\label{sec:conclusion}



\clearpage
\newpage
\bibliographystyle{named}
\bibliography{custom}

\clearpage
\newpage
\appendix
\section*{Appendix}

\subsection{Appendix: Problem Formulation}
\label{sec:problem}

\paragraph{Setting.}
We study data analysis in organizations where relevant information is distributed across \emph{heterogeneous} sources: some are directly queryable (e.g., SQL tables), while others are only available as \emph{visual documents} (e.g., scanned reports or screenshots containing tables). Existing analytics agents often perform well when the answer resides in a single structured source, but struggle to \emph{jointly reason} over structured data and table content embedded in images.

\paragraph{Data sources.}
Let $\mathcal{S}=\{S^{\texttt{sql}}, S^{\texttt{csv}}, S^{\texttt{nosql}}, S^{\texttt{txt}}, S^{\texttt{img}}\}$ denote the set of available sources:
(i) $S^{\texttt{sql}}$ is a relational database accessible via SQL;
(ii) $S^{\texttt{csv}}$ is a collection of CSV files;
(iii) $S^{\texttt{nosql}}$ is a semi-structured store (e.g., JSON documents);
(iv) $S^{\texttt{txt}}$ is a collection of unstructured text files;
(v) $S^{\texttt{img}}$ is a collection of images (or scanned PDFs) that contain \emph{tables} and other visually formatted records.
We emphasize that $S^{\texttt{img}}$ cannot be reliably queried without extracting table structure (rows/columns/headers) and cell values from pixels.

\paragraph{Input.}
The input to the system is:
\begin{equation}
X = \langle \mathcal{S},\, G \rangle,
\end{equation}
where $G$ is a natural-language analysis goal (a question or a higher-level objective), e.g.,
``Summarize the key cost drivers in 2023 and explain discrepancies between the finance database and the scanned invoices.''

\paragraph{Output.}
The system produces a response:
\begin{equation}
Y = \langle \texttt{Summary},\, \texttt{Insights},\, \texttt{Evidence} \rangle.
\end{equation}
\texttt{Summary} is a concise natural-language overview of the findings.
\texttt{Insights} is a list of specific, actionable findings (e.g., anomalies, trends, mismatches, root-cause hypotheses).
\texttt{Evidence} provides traceable support for each insight by pointing to the originating sources,
such as SQL rows/fields, CSV row indices, JSON paths, text spans, or image regions / extracted table cells.

\paragraph{Task objective (Unified Analysis).}
Given $X=\langle \mathcal{S}, G\rangle$, the goal is to generate $Y$ such that:
\begin{itemize}
    \item \textbf{Cross-source retrieval:} the system can access and use information from any subset of $\mathcal{S}$ required by $G$;
    \item \textbf{Cross-modal alignment:} data and column names mentioned or shown in $S^{\texttt{img}}$ can be matched to records in structured sources (e.g., SQL/CSV/NoSQL) when they refer to the same real-world item;
    \item \textbf{Joint reasoning for insights:} insights may require combining evidence from both structured data and visual tables;
    \item \textbf{Evidence grounding:} each reported insight is supported by explicit evidence links in \texttt{Evidence}, enabling verification.
\end{itemize}

\paragraph{Why this is non-trivial.}
The main difficulty is that $S^{\texttt{img}}$ is not directly queryable: answering $G$ may require (i) perception and table extraction from images, (ii) normalization of extracted values (units, formats, OCR noise), and (iii) reliable alignment/join operations between extracted visual records and structured tables. Our benchmark and agent are designed to evaluate and enable this end-to-end capability.

\subsection{Appendix: Human-in-the-Loop Construction and Verification}
\label{app:human_verification}

To ensure \textsc{DataCrossBench} reflects the complexity of real-world industrial scenarios while maintaining strict ground-truth accuracy, we implemented a two-stage \textit{Human-in-the-Loop (HITL)} pipeline. This process separates data construction (via AI-assisted synthesis) from quality assurance (via double-blind human verification).

\subsubsection{Stage 1: AI-Assisted Reverse-Synthesis}
\label{app:synthesis}

\begin{figure*}[t]
    \centering
    \begin{subfigure}[t]{0.48\linewidth}
        \centering
        \includegraphics[width=\linewidth]{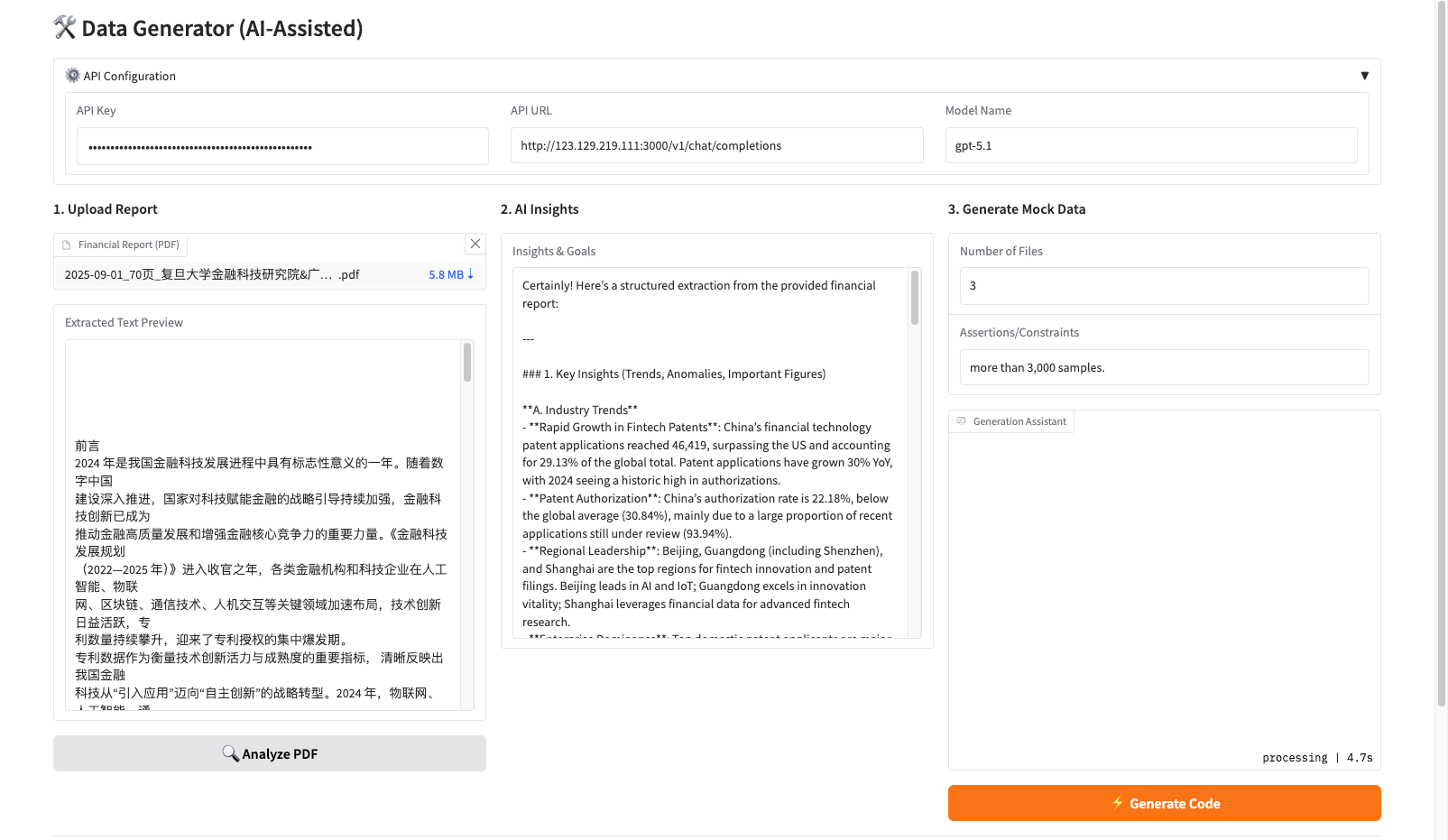}
        \caption{}
        \label{fig:plat3}
    \end{subfigure}
    \hfill
    \begin{subfigure}[t]{0.48\linewidth}
        \centering
        \includegraphics[width=\linewidth]{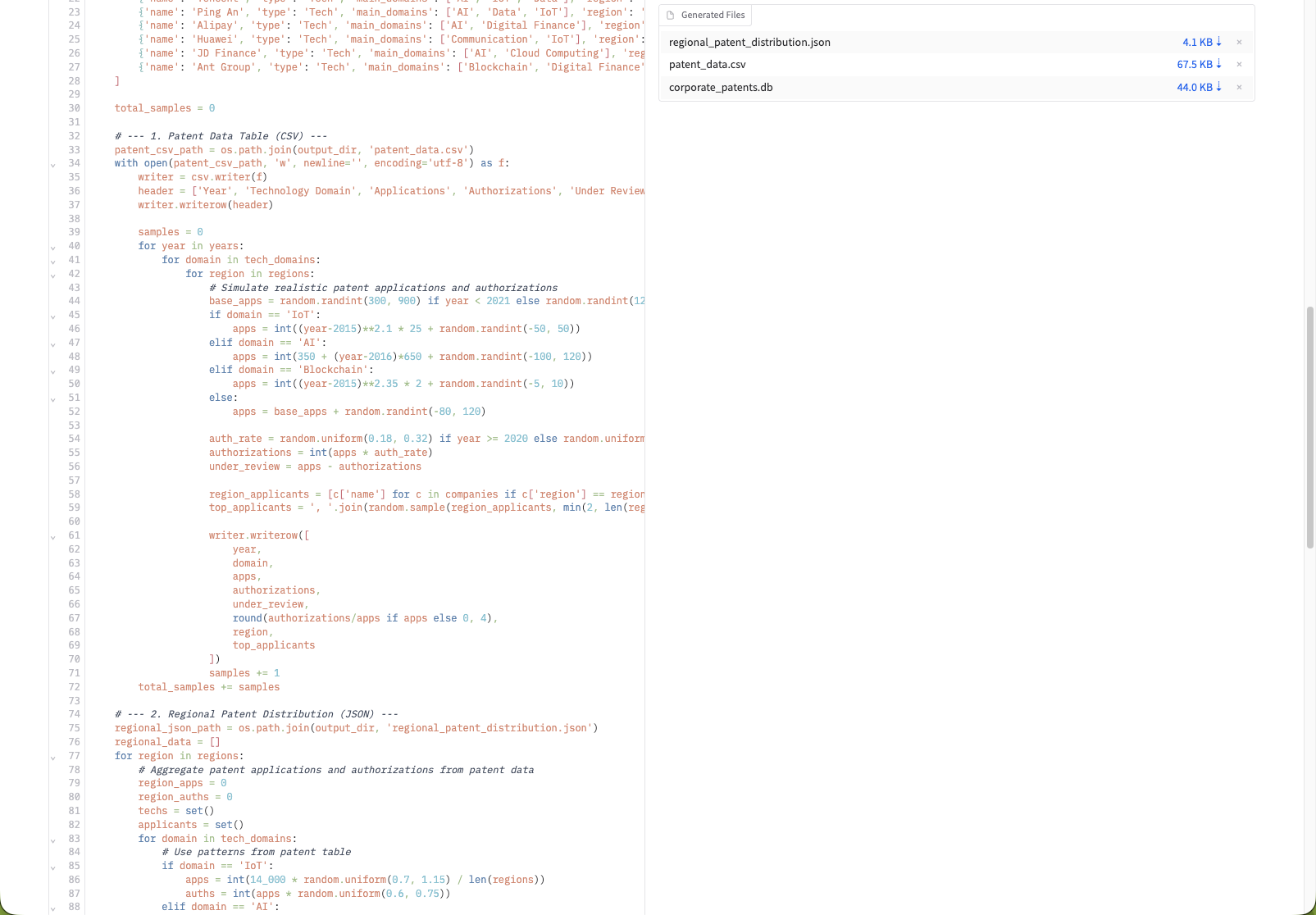}
        \caption{}
        \label{fig:plat4}
    \end{subfigure}
    \caption{\textbf{The AI-Assisted Data Generator Platform.} This interface facilitates the ``Reverse-Synthesis" pipeline, allowing experts to produce large-scale, heterogeneous datasets that adhere to specific logical constraints.}
    \label{fig:synthesis_platform}
\end{figure*}

Constructing heterogeneous datasets where structured files (SQL/CSV) perfectly correlate with unstructured visual documents (PDF/Images) is challenging. We developed a \textbf{Data Generator Platform} (Figure~\ref{fig:synthesis_platform}) to streamline this via a ``Reverse-Synthesis" workflow:

\begin{itemize}
    \item \textbf{Insight-Driven Generation (Fig.~\ref{fig:plat3}):} The workflow begins with a seed financial report. An LLM (gpt-4o) first extracts high-level insights and industry trends. Human experts then inject specific constraints (e.g., ``Generate a sales dataset with \>3,000 rows that contradicts the PDF summary by 5\%").
    \item \textbf{Code-Based Artifact Creation (Fig.~\ref{fig:plat4}):} Instead of asking the LLM to output raw data directly (which causes hallucinations), the platform generates executable Python/SQL scripts. These scripts synthesize the raw files and render the corresponding visual charts/tables, ensuring mathematical consistency between the ``zombie data" (images) and structured sources.
\end{itemize}

\subsubsection{Stage 2: Double-Blind Quality Verification}
\label{app:verification}

\begin{figure*}[t]
    \centering
    \begin{subfigure}[t]{0.48\linewidth}
        \centering
        \includegraphics[width=\linewidth]{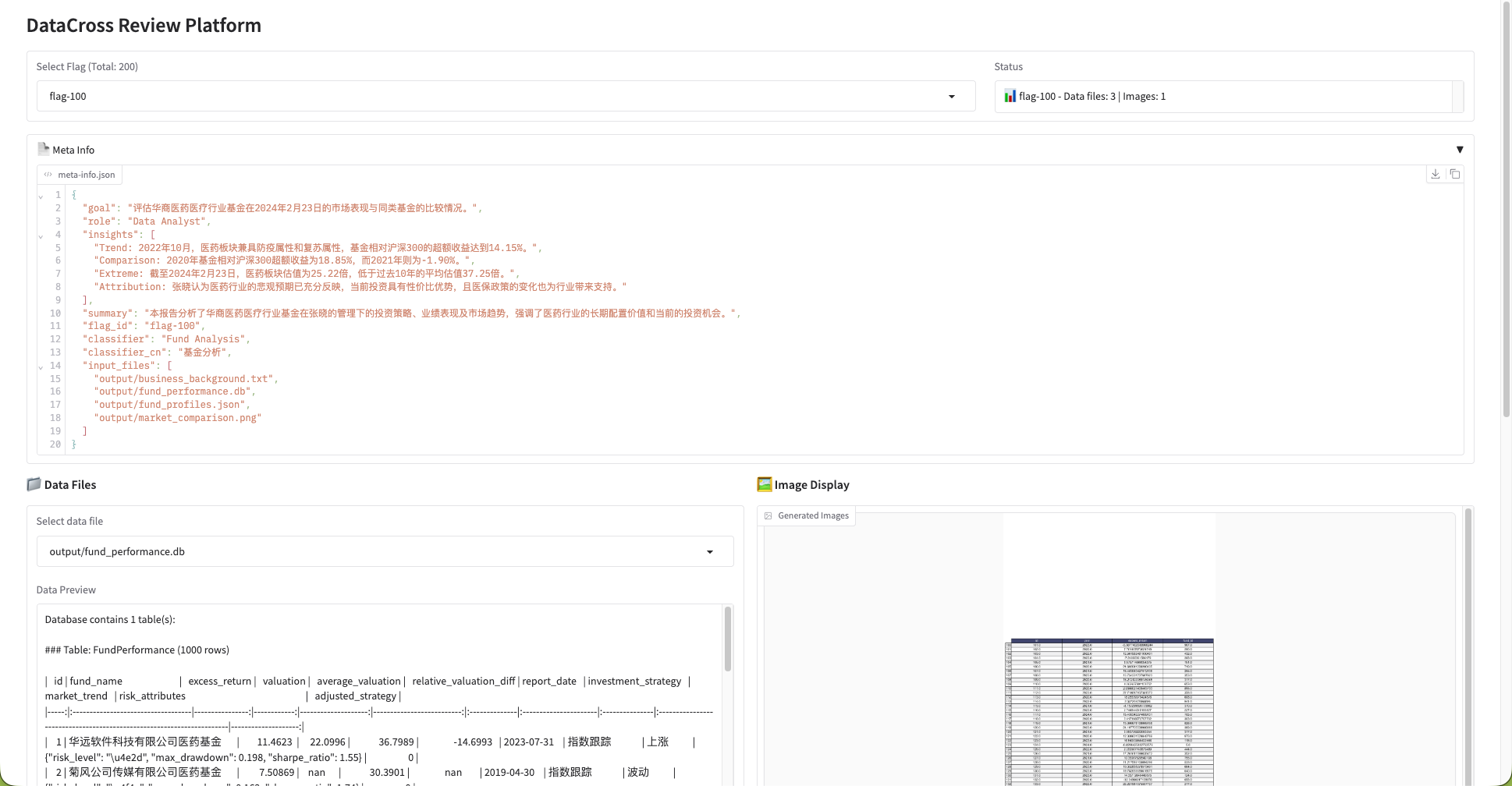}
        \caption{}
        \label{fig:plat1}
    \end{subfigure}
    \hfill
    \begin{subfigure}[t]{0.48\linewidth}
        \centering
        \includegraphics[width=\linewidth]{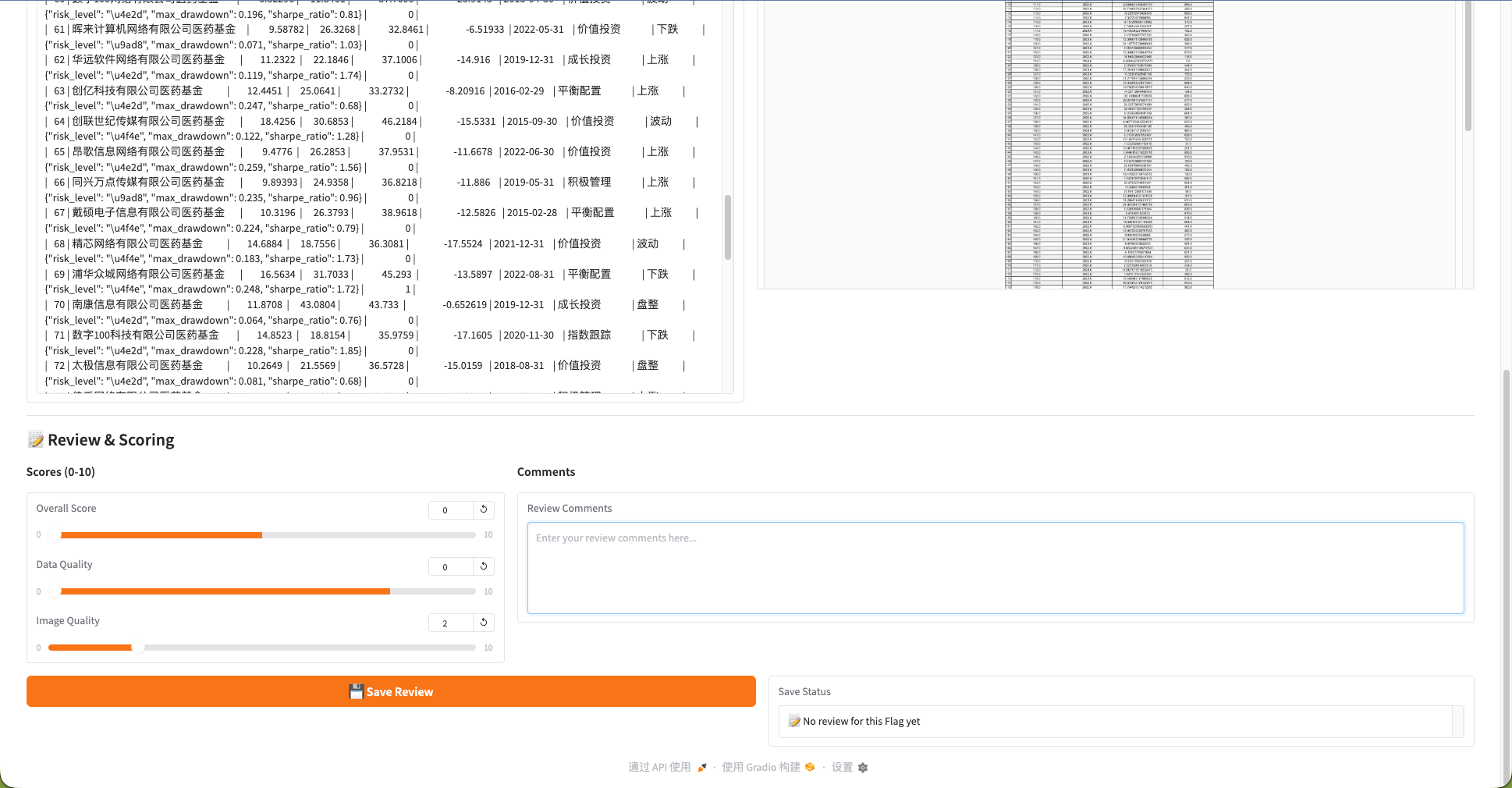}
        \caption{}
        \label{fig:plat2}
    \end{subfigure}
    \caption{\textbf{The DataCross Verification Platform.} This tool is used for the final quality audit, ensuring that the synthesized heterogeneous data supports the intended analysis goals without logical flaws or artifacts.}
    \label{fig:verification_platform}
\end{figure*}

All synthesized tasks undergo a rigorous review using the \textbf{DataCross Verification Platform} (Figure~\ref{fig:verification_platform}). This stage ensures that the generated ``zombie data" is legible and that the cross-source logic is sound.

\begin{itemize}
    \item \textbf{Unified Heterogeneous View (Fig.~\ref{fig:plat1}):} The interface provides a ``God's eye view" of the task. The left panel displays the ground-truth metadata and reasoning chains. The center and right panels allow annotators to inspect structured data (SQL/CSV previews) alongside the generated visual documents (charts/scanned tables) to verify cross-modal alignment.
    \item \textbf{Granular Scoring System (Fig.~\ref{fig:plat2}):} Annotators evaluate each task on a 0-10 scale across three dimensions: \textit{Overall Score}, \textit{Data Quality} (schema correctness), and \textit{Image Quality} (OCR legibility). The ``Save Review" module enforces a feedback loop; tasks scoring below 8.0 are flagged with specific comments (e.g., ``Image resolution too low for OCR") and returned to the synthesis stage for regeneration.
\end{itemize}

\subsection{Multi-Agent System Prompt Templates}
\label{appendix:prompts}

This appendix presents the full set of prompt templates used in the \textsc{DataCross} workflow, organized by execution stage.

\subsubsection{Stage 1: Initial Data Profiling}
\begin{lstlisting}[caption=PRELIMINARY\_EVAL\_PROMPT]
Role: Data Strategy Consultant
Global Goal: {global_goal}
Current Dataset Metadata (Schema & Stats):
{data_profile}

Task: 
Based ONLY on the schema and statistics provided, evaluate the potential relevance of this dataset to the Global Goal. Identify if this data contains core KPIs, key dimensions, or is likely just background noise.

Output Format:
Please wrap your evaluation in the following tags:
<relevance>1-10</relevance>
<reasoning>A brief explanation</reasoning>
<priority>High/Medium/Low</priority>
\end{lstlisting}

\subsubsection{Stage 2: Heuristic Exploration}

\begin{lstlisting}[caption=GET\_QUESTIONS\_TEMPLATE]
### Instruction:
GET_QUESTIONS_TEMPLATE = """
### Instruction:
Given the following context:
<context>{context}</context>
Given the following goal:
<goal>{goal}</goal>
Given the following schema:
<schema>{schema}</schema>

Instructions:
* Write a list of questions to be solved by the data scientists in your team to explore my data and reach my goal.
* Explore diverse aspects of the data, and ask questions that are relevant to my goal.
* You must ask the right questions to surface anything interesting (trends, anomalies, etc.)
* Make sure these can realistically be answered based on the data schema.
* The insights that your team will extract will be used to generate a report.
* Each question should only have one part, that is a single '?' at the end which only require a single answer.
* Do not number the questions.
* You can produce at most {max_questions} questions. Stop generation after that.
* Most importantly, each question must be enclosed within <question></question> tags. Refer to the example response below:

Example response:
<question>What is the average age of the customers?</question>
<question>What is the distribution of the customers based on their age?</question>

### Response:
"""

\end{lstlisting}

\begin{lstlisting}[caption=GENERATE\_CODE\_SINGLE\_TEMPLATE]
GENERATE_CODE_SINGLE_TEMPLATE = """
**Goal:** {goal}
**Question:** "{question}"
**Dataset Schema:**
{schema}

**File Path:**
The dataset is located at `{database_path}`.

---
**CRITICAL INSTRUCTIONS FOR WRITING PYTHON CODE:**

1.  **File Reading**:
    - You MUST load the file at `{database_path}` using the appropriate pandas function **based on its file extension** (e.g., `pd.read_csv()`, `pd.read_json()`).
    - **If reading a CSV file**: You MUST handle encoding errors. Use a `try-except` block. First, try `encoding='utf-8'`. If it fails, try `encoding='gbk'` or `encoding='latin1'`.
    - **Example for robust CSV reading**:
      ```python
      import pandas as pd
      file_path = '{database_path}' # This is the path
      try:
          df = pd.read_csv(file_path, encoding='utf-8')
      except UnicodeDecodeError:
          df = pd.read_csv(file_path, encoding='gbk')
      ```

2.  **CRITICAL: Column Names - USE EXACT NAMES FROM SCHEMA**:
    - **ALWAYS** use ONLY the exact column names shown in the schema above.
    - **DO NOT** assume nested column names like 'column_subfield'. The schema shows the actual column names.
    - **BEFORE** accessing any column, verify it exists: `if 'column_name' in df.columns:`
    - If you need data from a nested structure (e.g., dict/JSON in a column), use `.apply()`:
      ```python
      # Example: extracting from a nested column
      if 'stats' in df.columns and isinstance(df['stats'].iloc[0], dict):
          df['total_amount'] = df['stats'].apply(lambda x: x.get('total_transaction_amount', 0) if isinstance(x, dict) else 0)
      ```
    - **Example pattern for safe column access**:
      ```python
      # List actual columns from schema
      print("Available columns:", df.columns.tolist())
      # Then access only those that exist
      if 'column_name' in df.columns:
          value = df['column_name']
      else:
          print(f"Warning: 'column_name' not found. Available: {{df.columns.tolist()}}")
      ```

3.  **CRITICAL: Date/Time Columns**:
    - After loading the data, inspect the schema. If you see any columns that represent dates or times (e.g., 'date', 'timestamp'), you **MUST** convert them to datetime objects using `pd.to_datetime(df['column_name'], errors='coerce')`.
    - **DO NOT** attempt to use string methods like `.strftime()` on a column before converting it to datetime. All date operations **MUST** use the `.dt` accessor *after* this conversion.
    - Use `insight.tools.safe_datetime_parse()` for robust date parsing if standard methods fail.

4.  **Code Quality & Data Types**:
    - When creating a `pd.DataFrame` from a dictionary, ensure all arrays/lists have the same length to avoid `ValueError`.
    - Be mindful of data types. Do not assign string values to numeric columns or vice-versa, to avoid `FutureWarning`.
    - Use `insight.tools.safe_numeric_convert()` for converting mixed-type columns to numeric.

5.  **CRITICAL: Empty DataFrame Checks**:
    - After loading data, ALWAYS check if the DataFrame is empty: `if df.empty: print("Warning: Empty DataFrame")`
    - After filtering operations, check if the result is empty before proceeding.
    - Before aggregations (mean, sum, etc.), verify there is data to aggregate.
    - **Example pattern**:
      ```python
      filtered_df = df[df['column'] > threshold]
      if filtered_df.empty:
          print("No data matches the filter criteria")
          # Provide sensible defaults or skip the operation
      else:
          result = filtered_df['value'].mean()
      ```

6.  **Error Handling**:
    - Wrap critical operations in try-except blocks.
    - For column access, verify the column exists first: `if 'column_name' in df.columns:`
    - Handle KeyError, ValueError, and TypeError gracefully.

7.  **Output Generation**:
    - You **MUST** use the predefined functions from the `insight.tools` module to save all outputs.
    - **CRITICAL: Chinese Font Setup**: Before creating any plot, you MUST call `setup()` from `insight.tools` to ensure Chinese characters display correctly in plots. Add this line before any plotting code: `setup()`.
    - Generate one simple plot and save it as a `.jpg` file.
    - For the plot, save a statistics summary to `stat.json`.
    - Save the X and Y axis data (max 50 points) to `x_axis.json` and `y_axis.json` respectively.
    - Each JSON file must have "name", "description", and "value" fields. Ensure content is less than 4500 characters.
    - Call `insight.tools.fix_fnames()` at the very end of the script.

8.  **Code Structure**:
    - Start your code block with ```python and end it with ```.
    - Do not produce any text outside of this single Python code block.

**Available Tools:**
{function_docs}

---
Now, write the Python code to answer the question.

```python
"""
\end{lstlisting}

\begin{lstlisting}[caption=RETRY\_TEMPLATE]
RETRY_TEMPLATE = """You failed.

Instructions:
-------------
{initial_prompt}
-------------

Completion:
-------------
{prev_output}
-------------

Above, the Completion did not satisfy the constraints given in the Instructions.
Error:
-------------
{error}
-------------

Please try again. Do not apologize. Please only respond with an answer that satisfies the constraints laid out in the Instructions:

"""
\end{lstlisting}

\begin{lstlisting}[caption=FOLLOW\_UP\_TEMPLATE]
FOLLOW_UP_TEMPLATE = """
Hi, I require the services of your team to help me reach my goal.

<context>{context}</context>

<goal>{goal}</goal>

<schema>{schema}</schema>

<question>{question}</question>

<answer>{answer}</answer>

Instructions:
* Produce a list of follow up questions explore my data and reach my goal.
* Note that we have already answered <question> and have the answer at <answer>, do not include a question similar to the one above. 
* Explore diverse aspects of the data, and ask questions that are relevant to my goal.
* You must ask the right questions to surface anything interesting (trends, anomalies, etc.)
* Make sure these can realistically be answered based on the data schema.
* The insights that your team will extract will be used to generate a report.
* Each question that you produce must be enclosed in <question>content</question> tags.
* Each question should only have one part, that is a single '?' at the end which only require a single answer.
* Do not number the questions.
* You can produce at most {max_questions} questions.

"""
\end{lstlisting}

\begin{lstlisting}[caption=SELECT\_A\_QUESTION\_TEMPLATE]
SELECT_A_QUESTION_TEMPLATE = """
Hi, I require the services of your team to help me reach my goal.

<context>{context}</context>

<goal>{goal}</goal>

<prev_questions>{prev_questions_formatted}</prev_questions>

<followup_questions>{followup_questions_formatted}</followup_questions>

Instructions:
* Given a context and a goal, select one follow up question from the above list to explore after prev_question that will help me reach my goal.
* Do not select a question similar to the prev_questions above. 
* Output only the index of the question in your response inside <question_id></question_id> tag.
* The output questions id must be 0-indexed.
"""
\end{lstlisting}

\subsubsection{Stage 3: Key Source Identification}
\begin{lstlisting}[caption=FORMAL\_ANNOTATION\_PROMPT]
FORMAL_ANNOTATION_PROMPT = """
Role: Chief Data Scientist
Global Goal: {global_goal}
Given the following schema:
<schema>{schema}</schema>
Exploration Summary from the Agent's Deep-Dive:
{exploration_summary}

Task:
Perform a final assessment of this data's importance to the global objective.
Metrics:
- Information Richness (1-10): How deep and high-quality are the insights found?
- Theme Alignment (1-10): How directly does this support the Global Goal?

Decision Criteria:
- "Primary": Contains core metrics; can drive the main analysis.
- "Secondary": Provides context, auxiliary dimensions, or validation.

Output Format:
<richness>1-10</richness>
<alignment>1-10</alignment>
<label>Primary/Secondary</label>
<justification>Detailed reason</justification>
"""

\end{lstlisting}

\subsubsection{Stage 4: Crossover Pollinatio}
\begin{lstlisting}[caption=CROSS\_QUESTION\_PROMPT]
CROSS_QUESTION_PROMPT = """
Role: Cross-domain Analyst
Dataset A (Your Data) Summary: {my_summary}
Dataset B (Target Data) Summary: {other_summary}
Your Label: {my_label}

Task:
Generate analytical questions that require JOINING or COMPARING both datasets to find hidden patterns.
Constraint: 
- If your label is "Primary", generate 3 deep questions.
- If your label is "Secondary", generate 1 focused question.

Output Format:
Generate your questions, each enclosed in <question> tags.
Example: <question>Your question text (Rationale: ...)</question>
"""
\end{lstlisting}

\begin{lstlisting}[caption=ANNOTATION\_PROMPT\_TEMPLATE]
ANNOTATION_PROMPT_TEMPLATE = """
Role: Domain Expert & Critical Reviewer

You are: {annotator_name}
Your Domain Knowledge: {annotator_knowledge}
Your Data Schema: {annotator_schema}

You are reviewing analysis results from another agent.
Target Agent: {target_agent_name}
Target Agent's Analysis Insights: {target_insight}
Target Agent's Analysis Summary: {target_summary}

Task:
Provide critical comments or cross-domain insights based on your expertise.
Focus on:
1. Missing perspectives that your data might provide
2. Potential data quality issues
3. Alternative interpretations
4. Connections to broader business context

If you have no meaningful comments, respond with "no comment".
Otherwise, provide concise but insightful feedback.

Output Format:
<comment>Your critical feedback here</comment>
"""
\end{lstlisting}

\subsubsection{Stage 5: Final Summarize}
\begin{lstlisting}[caption=FINAL\_PROMPT\_TEMPLATE]
FINAL_PROMPT_TEMPLATE = """
Role: Senior Business Intelligence Analyst

Context from Complete Multi-Agent Analysis:
{full_context}

Task:
Synthesize all analyses, cross-dataset findings, and agent annotations into a comprehensive final report.

Your output should include:
1. Executive Summary (2-3 paragraphs)
2. Key Insights (bullet points, prioritized by importance)
3. Cross-dataset Discoveries
4. Limitations and Data Quality Notes
5. Recommended Next Steps

Format your response as a JSON object with the following structure:
{{
    "summary": "executive summary text here",
    "insights": ["insight 1", "insight 2", ...],
    "cross_dataset_discoveries": ["discovery 1", ...],
    "limitations": ["limitation 1", ...],
    "next_steps": ["recommendation 1", ...]
}}
"""
\end{lstlisting}

\end{document}